\documentclass[sigconf]{acmart}

\usepackage{subfigure}
\usepackage{todonotes}
\usepackage{url}
\usepackage{graphicx}
\usepackage{caption}
\usepackage{array,booktabs}
\newcolumntype{C}{@{}>{\kern\tabcolsep}l<{\kern\tabcolsep}}
\usepackage{colortbl}
\usepackage{xcolor}
\usepackage{amsfonts}
\usepackage{amsmath}
\usepackage{relsize}
\usepackage{balance}
\usepackage{array}

\usepackage{mathtools}
\DeclarePairedDelimiter\norm{\big\lVert}{\big\rVert}%
\makeatletter
\newcommand{\mylabel}[1]{\texttt{#1}\def\@currentlabel{\texttt{#1}}\label{#1}}

\newcommand{\mb}{ \mathbf}

\setcopyright{acmcopyright}

\acmDOI{xx.xxx/xxx_x}

\acmISBN{978-1-4503-6866-7/20/03}

\acmConference[SAC'20]{ACM SAC Conference}{March 30-April 3, 2020}{Brno, Czech Republic} 
\acmYear{2020}
\copyrightyear{2020}

\acmArticle{4}
\acmPrice{15.00}

\begin{document}

\copyrightyear{2020}
\acmYear{2020}
\setcopyright{acmcopyright}
\acmConference[SAC '20]{The 35th ACM/SIGAPP Symposium on Applied Computing}{March 30-April 3, 2020}{Brno, Czech Republic}
\acmBooktitle{The 35th ACM/SIGAPP Symposium on Applied Computing (SAC '20), March 30-April 3, 2020, Brno, Czech Republic}
\acmPrice{15.00}
\acmDOI{10.1145/3341105.3374013}
\acmISBN{978-1-4503-6866-7/20/03}

\title{Analysis of label noise in graph-based semi-supervised learning}

\author{Bruno Klaus de Aquino Afonso}
\affiliation{%
  \institution{Federal University of S\~ao Paulo}
  \city{S\~ao Jos\'e dos Campos} 
  \country{Brazil}
}
\email{isonettv@gmail.com}

\author{Lilian Berton}
\affiliation{%
  \institution{Federal University of S\~ao Paulo}
  \city{S\~ao Jos\'e dos Campos}
  \country{Brazil}}
\email{lberton@unifesp.br}

\renewcommand{\shortauthors}{Bruno Klaus de Aquino Afonso and Lilian Berton}
\begin{abstract}
In machine learning, one must acquire labels to help supervise a model that will be able to generalize to unseen data. However, the labeling process can be tedious, long, costly, and error-prone. It is often the case that most of our data is unlabeled. Semi-supervised learning (SSL)  alleviates that by making strong assumptions about the relation between the labels and the input data distribution. This paradigm has been successful in practice, but most SSL algorithms end up fully trusting the few available labels. In real life, both humans and automated systems are prone to mistakes; it is essential that our algorithms are able to work with labels that are both few and also unreliable. Our work aims to perform an extensive empirical evaluation of existing graph-based semi-supervised algorithms, like Gaussian Fields and Harmonic Functions, Local and Global Consistency, Laplacian Eigenmaps, Graph Transduction Through Alternating Minimization. To do that, we compare the accuracy of classifiers while varying the amount of labeled data and label noise for many different samples. Our results show that, if the dataset is consistent with SSL assumptions, we are able to detect the noisiest instances, although this gets harder when the number of available labels decreases. Also, the Laplacian Eigenmaps algorithm performed better than label propagation when the data came from high-dimensional clusters.
\end{abstract}

%
%



\maketitle

\section{Introduction} \label{sec:introduction}

\textit{Machine learning} (ML) is the subfield of Computer Science that aims to make a computer learn some task from experience \cite{Mitchell97}. As such, it has been used extensively to extract meaningful knowledge from data. In order to learn something from the data, many of the ML approaches require some form of annotation. These annotations referred hereafter as \textit{labels}, are useful in many ways. Often, it is the attribute that we want to predict in future data when considering a classification or regression task.  

In most real-life scenarios, the assumption that all labels are available is not good enough \cite{Chapelle_etal_2010}.  Labeling 
data, be it by specialists or crowd-sourcing, often consumes too much time and money. The process is also tedious and error-prone. As a result, it is desirable to have good classification results using as little labels as possible. \textit{Semi-Supervised Learning} (SSL) is a paradigm suited for when only a small subset of the data is labeled. The large amounts of easy-to-obtain, unlabeled data serve to guide the labeled information by modifying or re-prioritizing hypotheses \cite{Zhu_2005}. This approach has been applied to a multitude of tasks, such as community detection \cite{xie2013labelrank}, computer vision \cite{miyato2018virtual}, image processing \cite{BERTON2017,Katunda2019}, drug-protein interaction  prediction \cite{xia2010semi}, sentiment analysis \cite{hamilton2016inducing} and  word sense disambiguation\cite{yuan2016semi}.

Although the traditional semi-supervised algorithms are designed to make most of the few labels they have available, they are usually not robust to label noise. In the real world, humans get tired and automated systems get fooled. As such, it would be wise to not fully trust the labeling process, and consider the labels to be \textit{weak}, i.e. unreliable to some extent.
 If the semi-supervised algorithm believes blindly on the labels it is given, any labeling error could be propagated and affect classifier performance significantly. Noise can be defined as anything obscuring the relationship between the class and the features \cite{Hickey_1996}. The three major sources of noise \cite{Hickey_1996} are the insufficiency of the description schema, corruption of the input features (\textit{feature noise} or \textit{attribute noise}), and misclassification 
of training examples (\textit{label noise} or \textit{class noise}).  However, for real-world datasets, it is difficult to measure the insufficiency of the description schema \cite{Zhu2004}, so usually, the other two sources are considered.

There are three ways of dealing with label noise. First, one could use \textit{label noise-robust models} that do not explicitly handle label noise. They can only hope that the usual overfitting avoidance mechanisms will also be useful to lessen the impact of label noise. This may be a consequence, i.e., of the chosen loss function. It has been shown that the squared loss is tolerant only to uniform noise \cite{manwani2013noise}.  The second type of approach would be the use of \textit{filters} to eliminate the noise from the training set before using it for learning a classifier. This can be done by either removing instances with noisy labels or trying to correct them \cite{teng2001comparison}. Thirdly, \textit{label noise-tolerant} methods do consider label noise directly. In particular, supervised and semi-supervised learning algorithms can be modified to be tolerant to noise. A comprehensive label noise overview is given by \cite{Frenay2014}.

The scarcity and the unreliability of labels have been individually well studied. Despite this, there hasn't been much research that focuses on methods that tackle both problems at once. This work aims to analyze the label noise in semi-supervised learning, especially in graph-based methods that are most common in literature \cite{zhu2002,zhou2004learning,Zhu_2003,Berton2018,Vega_Oliveros_2014,Berton_2015}. These methods represent each instance as a vertex in a graph, and neighboring vertices are connected by edges weighted by some similarity metric. This results in a smoothness criterion that discourages different predictions in vertices with strong links to each other, label propagation allows us to spread the known labels to the unlabeled vertices through the graph structure.

The main objective of this work is empirically analyze different graph-based SSL algorithms in order to measure their accuracy in the presence of label noise and whether label noise-robust algorithms exist. This includes varying the amount of labeled data, as well as the amount of noise within those labels, using different seeds to determine the sampling and noise process.
In this work, we aim to answer:
\begin{itemize}
    \item Are there any graph-based semi-supervised algorithms that are robust to label noise?
    \item Many algorithms have hyperparameters dictating the importance of fitting the classifying function to the observed labeled data. To what extent does the tuning of those parameters reduce the effect of label noise?
    \item Which GSSL algorithms have predictions with lower variance when label noise is present?
    \item Which assumptions must be made about a dataset in order to detect noisy instances in a semi-supervised manner?
\end{itemize}    

The remaining of this work is organized as follows. Section \ref{related_work} presents some graph-based SSL approaches and their relation with label noise. Section \ref{materials_methods} presents the main SSL concepts related to this work, the algorithms and datasets used for the experiments. Section \ref{results} presents the results and finally, Section \ref{sec:conlusion-and-future-works} presents the final remarks.

\section{Related work} \label{related_work}

Graph-based methods have been a staple of semi-supervised learning for some time.  Many of the graph-based semi-supervised methods may be formulated as a convex optimization method. This means that they are guaranteed to converge to the global optimum, as opposed to other semi-supervised algorithms. They also do not require to comply with a parameterized decision rule. Some of the resulting algorithms follow a simple iteration rule, being easy to implement. Finally, they may also be interpreted in different views, such as a random walk, a minimization of a quadratic criterion, or even as the solution to the heat equation \cite{Zhu_2003}.

One of the earliest label propagation algorithms, named \emph{Gaussian Fields and Harmonic Functions} (GFHF) \cite{Zhu_2003} did not address labels unreliability. Another classic graph-based algorithm, named \emph{Local and Global Consistency} (LGC) \cite{zhou2004learning}\label{alg:LGC}, takes a step to address this issue. Whereas GFHF fully prioritizes the fitness criterion $\norm{\widehat{Y_l} - Y_l}^2$ over the smoothness criterion, LGC introduces a hyperparameter $\lambda$ to regulate the trade-off. There is a similar trade-off in the \emph{Adsorption} algorithm \cite{Baluja2008}, where one can lower the injection probability if the labels are not to be trusted. Another version of this type of regularization is due to \cite{belkin2004regularization}. This algorithm has been used in applications to address noisy labels, e.g. in image and video annotation \cite{gao2015optimal}. The regularization of \emph{Laplacian Eigenmaps} (LE) \cite{Belkin2003} outright restricts the class of functions used. Using a combination of the first $p$ smoothest eigenfunctions could be useful, especially if the less smooth functions would be used to fit noisy instances.

 The $\ell_1$-norm has been employed for \textit{fine-grained labeling of large shape collections} \cite{huang2013fine}, that is, dividing objects (in this instance, 3D models) of the same category into sub-categories. 
 \textit{Approximate kNN-SGSSL with noisy label handling}\cite{tang2011image} uses $\ell_1$-norm on both graph construction and cost minimization.
 \textit{Large-Scale sparse coding} (LSSC) \cite{lu2015noise} uses the $\ell_1$-norm to transform noise-robust semi-supervised learning into a generalized sparse coding problem.
 {Semi-Supervised learning under Inadequate and Incorrect supervision} \cite{gong2017learning} applies to the unnormalized Laplacian $\mb{L}$ ideas similar to the ones in LSSC.
 Semi-supervised learning with noise can also be seen as a graph-signal restoration problem, as in \cite{mao2016image}, which uses a \textit{generalized graph smoothness prior} to learn an image classifier given noisy labels.

\section{Materials and methods} \label{materials_methods}
\emph{Machine learning} (ML) always uses some representation of the data. Usually, this representation comes as a collection of $n$ vectors with $d$ dimensions:
\begin{align}
\mathcal{X} = \{ \mb{x}_1,\ldots,\mb{x}_n \} \end{align}
Each vector is called an \textit{instance}, and each of its components an \textit{attribute}, or \textit{feature}. It is assumed that all instances were drawn independently and identically distributed from some probability distribution $P(\mb{x})$.
Most often, the objective of a model is to predict some attributes for novel instances. This attribute is hereafter referred to as a \textit{label} for each instance. In a \textit{classification} task, the label may only have one of the finitely possible values, each called a \textit{class}.  

\textit{Semi-Supervised learning} (SSL) is a paradigm that addresses the situation where only the labels of a few  instances are available, that is
\begin{equation}
\mathcal{Y} = \{ y_1,\ldots,y_l \}, \;\; l \ll n = l + u
\end{equation}
The input instances are divided accordingly:

\begin{equation}
\mathcal{X} = X_l \cup X_u = \{ \mb{x}_1,\ldots,\mb{x}_l \} \cup
 \{ \mb{x}_{(l+1)},\ldots,\mb{x}_{(l+u)} \} 
\end{equation}

The key principle behind any SSL classifier is that the data distribution $P(\mb{x})$ can tell us something about the conditional distribution of the labels, i.e. $P(y\!\mid\!\mb{x})$. This notion of usefulness can be made more precise. To do so, some assumptions are employed by the algorithms, such as the Smoothness, Cluster and Manifold assumptions \cite{Chapelle_etal_2010}.

\begin{figure}[!h]
    \centering
        \includegraphics[scale=0.5]{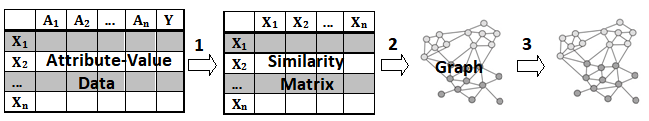}
        \caption{Steps to apply graph-based SSL.}
    \label{fig:ssl}
\end{figure}

The three main steps taken in \emph{graph-based semi-supervised learning} (GSSL) algorithms \cite{Jebara_etall_2009,Berton2018} are presented in Figure \ref{fig:ssl}.
First, given a dataset $X$ in attribute-value format, we need to calculate similarity among the examples; then, a graph construction method needs to be employed in order to generate a graph $G$; finally, this information is provided to a classifier, so that it may return the predicted labels for all instances. This classifier may use, e.g., a label propagation algorithm to spread the known labels to the unlabeled examples. 

The purpose of a measure of similarity is to compare two lists of numbers (i.e. vectors) and compute a single number that evaluates their similarity. The basis of many measures of similarity is Euclidean distance. 

A weighted graph emerges as a natural approximation to the intrinsic geometric structure of the data. Vertices are the union of labeled instances $X_l = \{\mb{x}1,\ldots, \mb{x}l \}$ and unlabeled instances $X_u = \{\mb{x}_{l+1},\ldots, \mb{x}_{l+u} \}$. 
It can be represented as $G = (V, E, W)$ , where $V = \{v_1, v_2, \ldots, v_n\}$ is a set of $n$ vertices and $E = \{e_1, e_2, \ldots, e_m\}$ is a set of $m$ edges connecting pair of vertices $v_i$ and $v_j$. The edges can have weights represented by the \textit{affinity matrix} $W: V \times V \rightarrow \mathbb{R}$ that is a measure of similarity between data instances. 
 
The most common methods for graph construction is the \emph{k-nearest neighbor} (KNN) graph in which two vertices $v_i$ and $v_j$ are connected by an edge, if the distance between $v_i$ and $v_j$ is among the k-th smallest distances from $v_i$ to other objects from $G$. 

One of the most pivotal elements of graph-based SSL, the \textit{graph Laplacian} is defined as
\begin{equation}
    \label{def:graph_lapl}
L = D - W
\end{equation}
where D, called the \textit{degree matrix}, is a diagonal matrix with entries
\begin{equation}
D_{ii} = \sum_{j: (i,j) \in E}W_{ij}
\end{equation}
The graph laplacian has the same properties as what you'd expect from a discrete analogue of the Laplacian-Beltrami operator.

Given a dataset $X = \{X_l \cup X_u\}$ represented as a graph $G = (V, E, W)$, the inference task is to diffuse the known labels $Y_l$ to the unlabeled vertices $X_u$ by estimating their labels $Y_u$. Several methods for propagating labels have been proposed. 
Many label propagation algorithms iteratively compute a distribution of labels on the vertices of the graph to maximize the consistency with the clustering and manifold assumption. Some well-known label propagation algorithms are considered here and described in the next section.

\subsection{Graph-based SSL algorithms (GSSL)}
In \emph{Gaussian Fields and Harmonic Functions} (GFHF) algorithm \cite{zhu2002} the label propagation occurs via the iteration of the following:
\begin{align}
\widehat{Y}^{(t+1)} &= P \, \widehat{Y}^{(t)} ; \;\;\; \widehat{Y}_l^{(t+1)} = Y_l
\end{align}
Here, $P=D^{-1}W$ is the row-normalized weight matrix and $\widehat{Y}^{(0)} = Y = [Y_l,Y_u]^{\top}$ is the matrix representing the initial labeling.

\emph{Local and global consistency} (LGC) algorithm \cite{Zhou_etall_2004} is similar but introduces a new parameter $\alpha$. The initialization is the same, and the iterative update is: 
\begin{equation}
\widehat{Y}^{(t+1)} = \alpha (D^{-1/2}WD^{-1/2}) + (1-\alpha)\widehat{Y}^{(0)}
\end{equation}

\emph{Laplacian Eigenmaps} (LE) algorithm \cite{Belkin2003} uses a different approach than previous algorithms. Namely, the smoothness criterion is enforced by restricting the solution to a combination of smooth eigenfunctions. The LE algorithm consists of an unsupervised step and a supervised step. In the first step, we consider the minimization of the smoothness criterion  $\widehat{Y}^{\top} L \widehat{Y}$. The second step then uses the labeled information to derive the best coefficients $a_1, \ldots , a_p$ to be used as a label-predicting linear combination of smooth eigenfunctions. This is achieved by minimizing the error
\begin{equation}
    Err(a_1, \ldots, a_p) = 
    \sum^l_{i=1}\left( Y_i - \sum_{j=1}^p a_j \mb{f}_j(i) \right)^2
\end{equation}

Both GFHF and LGC fit in a univariate regularization framework, where  $\mb{F}$ is the only variable considered. Thus, they do not try to explicitly correct the initial labels in $\mb{Y}$. This is addressed by the \emph{Graph Transduction via Alternating Maximization} (GTAM) algorithm \cite{wang2008graph}, as it minimizes the following bivariate criterion:
\begin{equation}
    Q(\mb{F},\mb{Y}) = \frac12 tr \left(\mb{F}^{\top}\mathcal{L}\mb{F} + \lambda (\mb{F} - \widetilde{\mb{Y}})^{\top}
    (\mb{F} - \widetilde{\mb{Y}}) \right) 
\end{equation}
Here, $\widetilde{\mb{Y}}$ is a modified version of the current $\mb{Y}$, such that  columns sum up to 1 and nodes with high degree values are given more weight:
\begin{equation}
    \widetilde{\mb{Y}}_{ij} = 
    \frac{D_{ii}}{\sum_{1\leq k \leq n}D_{kk}Y_{kj}} Y_{ij}
\end{equation}

\subsection{Datasets}
\begin{figure*}[!htb]
    \centering
        \subfigure[g241c.]{\label{fig:g241c}
        \includegraphics[width= 0.25\textwidth]{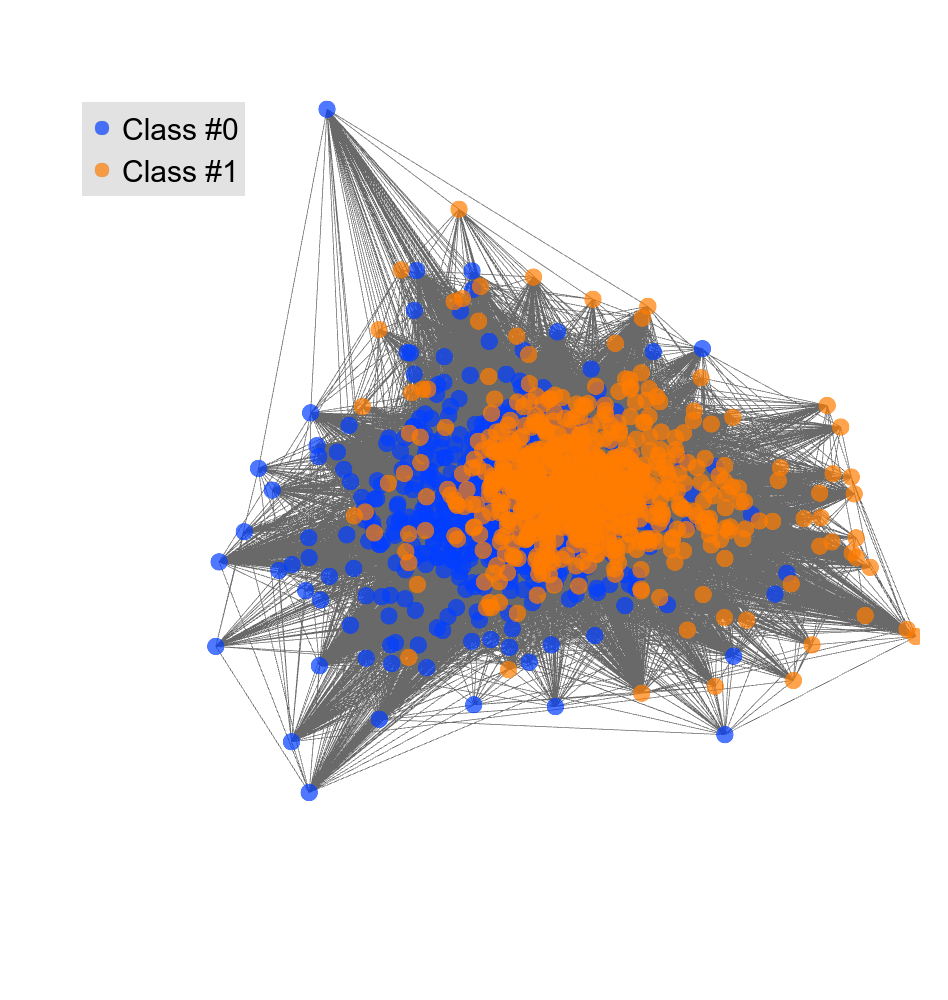}}
        \subfigure[g241n.]{\label{fig:g241n}
        \includegraphics[width= 0.25\textwidth]{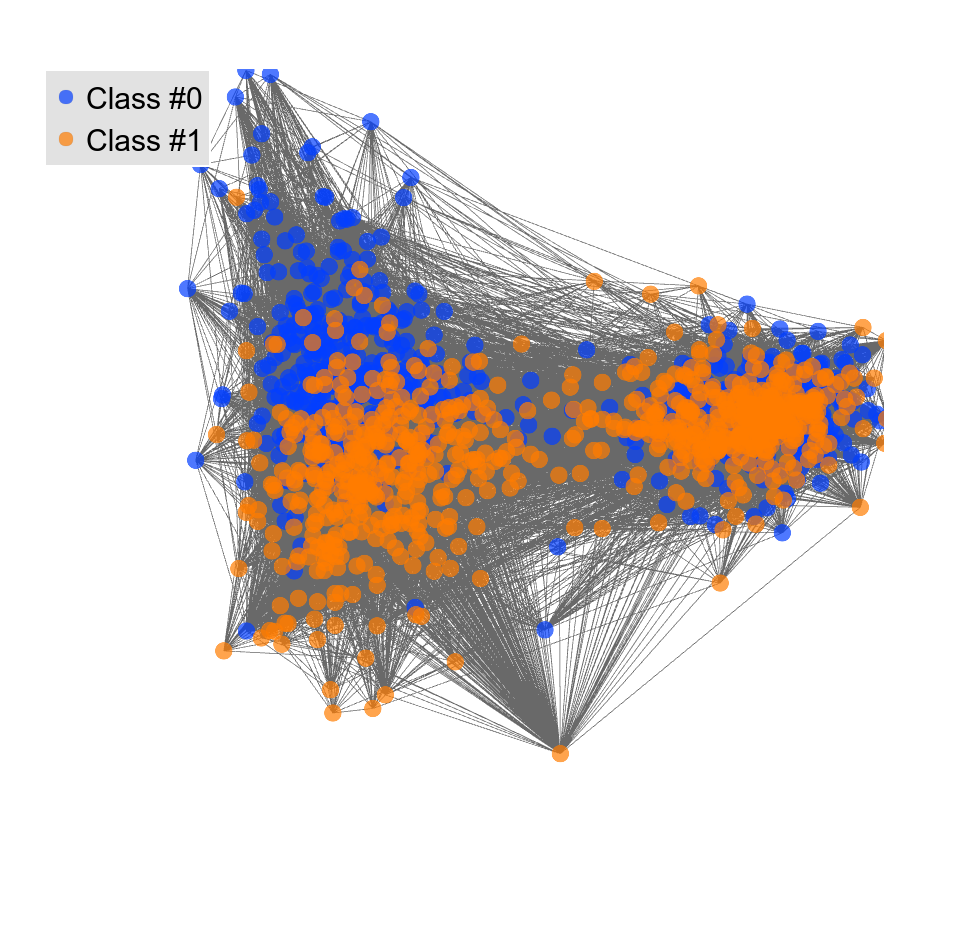}}
         \subfigure[Digit1.]{\label{fig:digit1}
        \includegraphics[width= 0.22\textwidth]{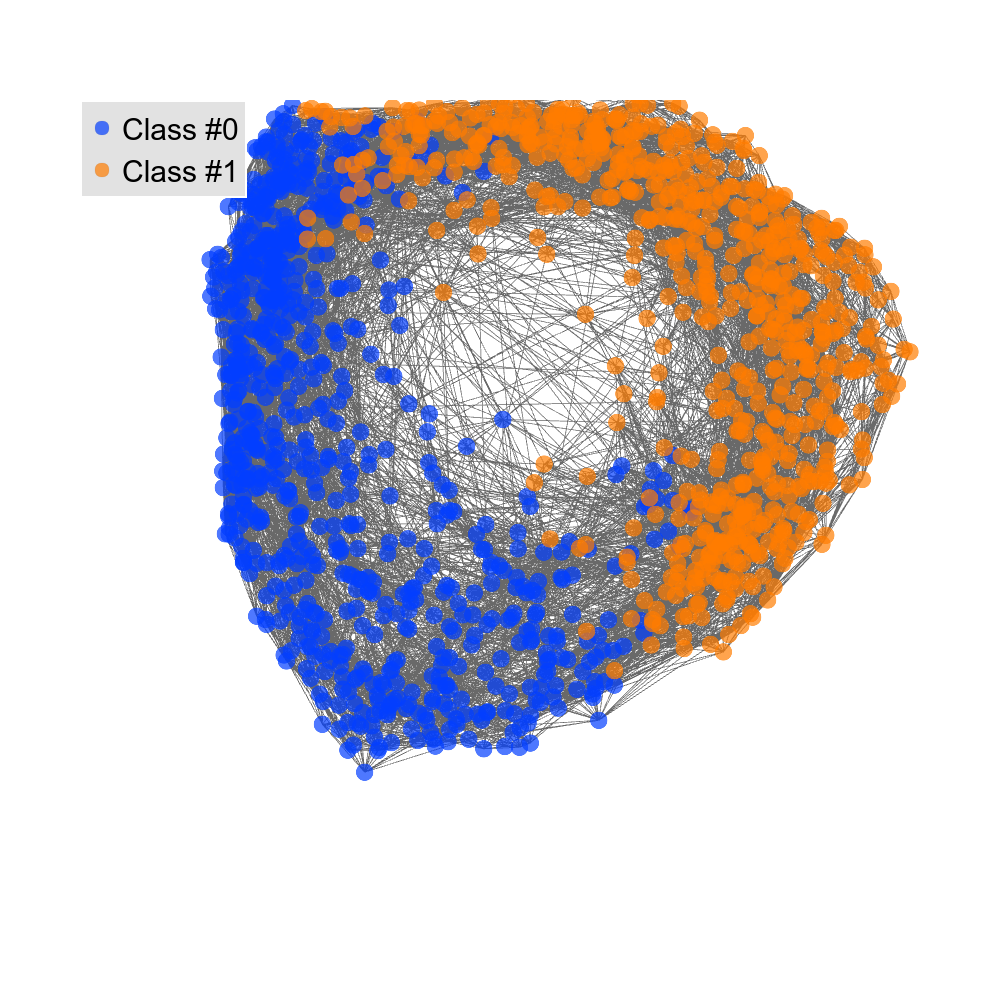}}
         \subfigure[COIL2.]{\label{fig:coil2}
        \includegraphics[width= 0.25\textwidth]{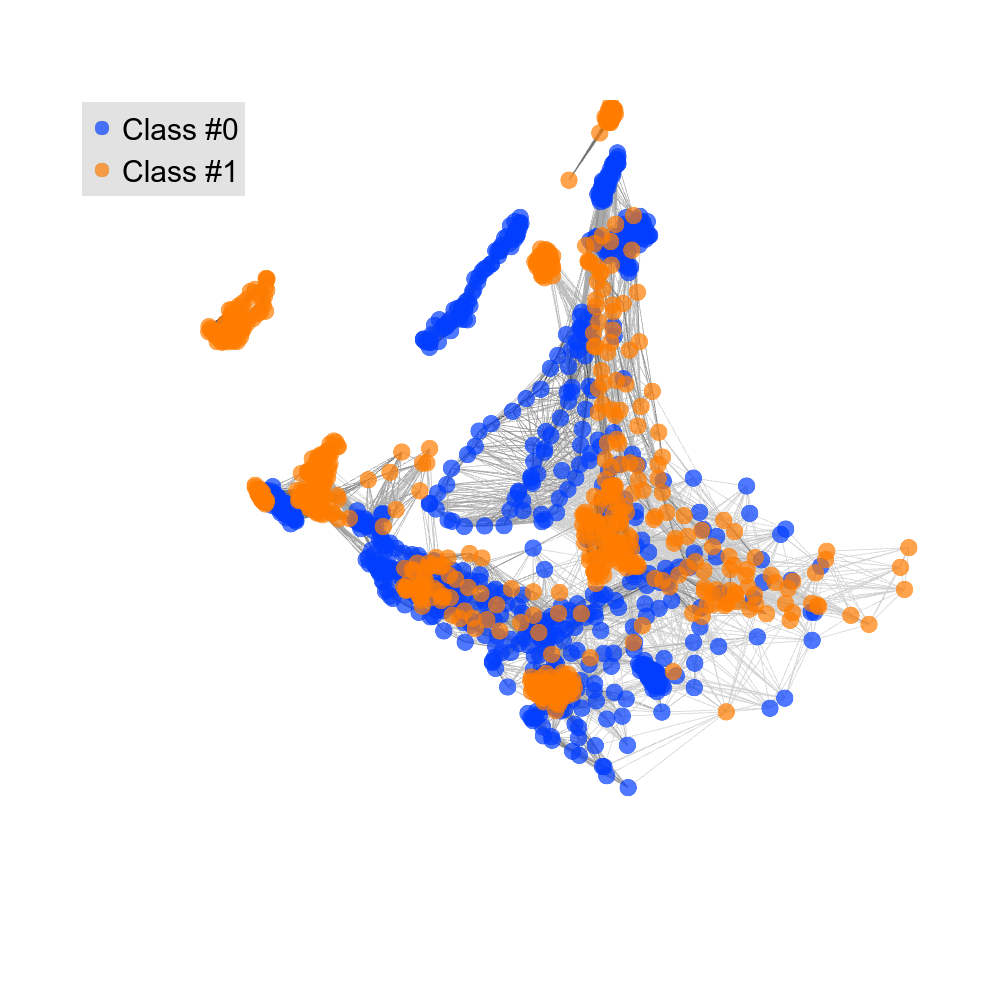}}
    \caption{Locally linear embedding (LLE) of the Digit1, COIL2, g241c and g241n datasets. The number of neighbors parameters for the LLE for g241c was set to 15, for g241n was set to 40, for Digit1 was set to 15 and for COIL2 was set to 150.}
    \label{fig:dataset}
\end{figure*}

Graph-based semi-supervised learning (GSSL) relies on the manifold assumption, as well as the semi-supervised smoothness assumption and cluster assumption. These sorts of assumptions are valid in many datasets, but not all. With that in mind, we chose our datasets such that there is diversity with respect to the extent that each assumption is being classified. All datasets have the same number of dimensions (241) and points (1500). More details are presented below. For every one of them, an illustration is provided (Figure \ref{fig:dataset}).  Whenever dataset with more than two dimensions, we perform a Locally Linear Embedding (LLE) \cite{roweis2000nonlinear}, as it makes use of similar assumptions and can give us some insight about the output of GSSL classifiers.

\paragraph{g241c and g241n}
The g241c dataset \cite{Chapelle_etal_2010}  was created such that the cluster assumption holds, but the manifold assumption does not.  Input is drawn from two 241-dimensional Gaussians, and the label of an instance corresponds to gaussian it was drawn from. The g241n dataset is based on 4 Gaussians: A1, A2, B1, B2. Say that the first two Gaussians correspond to class A, and the latter two to class B. There is more overlap between A1 and B1 or A2 and B2 than between A1 and A2 or B1 and B2. There is much inter-class overlap, and little intra-class overlap, leading to a misleading cluster structure. These datasets are illustrated in Figures \ref{fig:g241c} and \ref{fig:g241n}, respectively.

\paragraph{Digit1}
The Digit1 dataset  \cite{Chapelle_etal_2010} consists of artificially generated images of the digit 1. This satisfies the manifold assumption, as the images were produced according to the specification of 5 parameters:
two for translation, one for rotation, one for line thickness and one for the length of the small line at the bottom. Downsampling and omission of certain pixels reduce the 16 by 16 image to a 241-dimensional instance. This dataset does not show an obvious cluster structure. It is illustrated in Figure \ref{fig:digit1}.

\paragraph{COIL2}
The COIL2 dataset \cite{Chapelle_etal_2010} is derived from the the Columbia object image library (COIL-100)\cite{nene1996columbia}, which contains  a set of colored images of 100 different objects. COIL2 has binary labels and uses 24 out of the 100 objects. Once again, the dataset is downsampled enough to be reduced to 241 dimensions. Each object has pictures taken from different angles, in steps of 5 degrees. Thus, we should expect there to be a path in the graph passing through each consecutive angle and wrap around, ending up at the first picture for the object. This dataset has some well-separated clusters, which is good for SSL. However, when there are very few labels, it could happen that there simply isn't any label for some small cluster. When there is noise, the only available label could be unreliable. Moreover, there is a bit of overlap in some regions. Therefore, this is a harder dataset than Digit1. This dataset is illustrated in Figure \ref{fig:coil2}. 

\section{Results} \label{results}
This section presents the experiment configuration and the results of applying four graph-based SSL algorithms on the four datasets described previously, varying the percentage of labels and noise. 

\subsection{Experiment configuration}
\label{subsec:prop_exp_config}
The main challenge when performing this sort of comparison work is dealing with the combinatorial explosion that leads to a huge amount of configurations. To make this more evident, let us consider the parameters that must be fixed for each configuration for a set experiment. Any configuration can be divided into 6 subconfigurations:
\begin{itemize}
\item \textit{Random}: Random seed for sampling the dataset and determining labels to be corrupted with noise. We chose to have 20 such seeds.
\item \textit{Data}: The chosen dataset.
\item \textit{Label}: The percent of data instances that are labeled. This can be interpreted as setting the value for the fraction $\frac{l}{l+u}$, where $l$ and $u$ are the numbers of labeled and unlabeled examples, respectively.
\item \textit{Noise}: The noise process the label is subject to. For each class, we simply change a set percentage of its labels to the other class.
\item \textit{AffMat}: The chosen construction method that maps the input features to an affinity matrix. Includes hyperparameters such as 
$k$ for a mutual KNN graph.
\item \textit{Alg}: The chosen GSSL algorithm, including the setting of any hyperparameters. These are
\begin{itemize}
    \item GFHF: Gaussian Fields and Harmonic Functions
    \item LGC: Local and Global Consistency
    \item LE: Laplacian Eigenmaps 
    \item GTAM: Graph Transduction Through Alternating Minimization
\end{itemize}{}
\end{itemize}
A thorough comparison would require a number of configurations equal to: 
\begin{equation} 20*|\text{\texttt{Data}}|
    *
    |\text{\texttt{Label}}|
    *
    |\text{\texttt{Noise}}|
    *
    |\text{\texttt{AffMat}}|
    *
    |\text{\texttt{Alg}}|
\end{equation}
For this very reason, we chose to use smaller datasets. 

\subsection{Experiment results}
For the experiments, we evaluate the robustness of some GSSL algorithms (GFHF, LGC, LE, GTAM) on the  \textbf{Digit1}, \textbf{COIL2}, \textbf{g241n} and \textbf{g241c} datasets. We used a fixed configuration for the affinity matrix, which used $k=15$ neighbors for the mutual KNN graph, and $W_{ij} = 1$ if $x_i$,$x_j$ are neighbors. The reasoning behind this is that, when label noise is present, one should not give too much weight to the nearest instances, as they could be corrupted. It is worth noting that many of the underlying differences between GSSL algorithms are related to how they treat vertices with a higher degree (or some measure of centrality), and those sorts of differences are mostly nullified in this setting.

\paragraph{Digit1}
The results for Digit1 are presented on Table \ref{tab:digit1} and the following are the main observations:
\begin{itemize}
    \item LE using the top $20\%$ of eigenfunctions performed well on this dataset when there were very few labels. 
    \item LE, as opposed to the other algorithms, actually got better results when there were fewer labels. We believe that due to the simple structure of the Digit1 dataset, using the first few eigenfunctions to build a linear classifier should in practice identify and label the two observed regions. It could be that the first few eigenfunctions are preferred when there are fewer labels to fit, as they provide a simpler hypothesis that relies more on the unlabeled structure of the data. This is further addressed in section \ref{sec:disc}.
    \item The GTAM algorithm with $\mu=0.0101$ seemed to resist noise very well whenever there was a considerable amount of labels. Under extremely heavy noise (35\%), it has a high variance. 
    
    \item Let us consider the accuracy decrease due to 20\% noisy labels when 10\% of instances are labeled.
    \begin{itemize}
        \item GFHF: 5.1 \%
        \item LGC with $\alpha=0.9$: 5.5\%;
        \item LGC with $\alpha=0.1$: 14.4\%;
        \item GTAM with $\mu$ = 99: 6.5\%
        \item GTAM with $\mu$ = 0.0101: 0.9\%
        \item LE: 9.8\%
    \end{itemize}
    For the 1\% labeled case, we omit GTAM, as it had too much variance. For the remaining algorithms, the decrease was: %
    \begin{itemize}
         \item GFHF: 14.5\%
        \item LGC with $\alpha=0.9$: 11.8\%;
        \item LGC with $\alpha=0.1$: 16.1\%;
        \item LE: 7.7\% 
    \end{itemize}
    
    \item LE tolerated noise very well up to 20\% of corrupted labels. It was also mostly unaffected when there were very few labels. We believe that the number of eigenfunctions used (20\% of the total) is a bit excessive, as the first few eigenfunctions could be enough to separate the classes. If LE successfully identifies the regions corresponding to each cluster, and each cluster corresponds to a class, then a within-cluster majority vote would possibly be the best way to deal with label noise.  
    \item The GTAM algorithm had a high variance when subject to extremely heavy noise (35\%) or very few labels (1\%). It could be that it resists heavy noise well in most cases, but does extremely poorly in a few.
\end{itemize}

\begin{table*}[!htb]
\scriptsize
\begin{tabular}{@{}lllllllll@{}}
\toprule
Algorithm & $\alpha$ & $\mu$  & p   & Noise & Acc. (10\% labeled)          & Acc. (5\% labeled)           & Acc. (2.5\% labeled)         & Acc. (1\% labeled)           \\ \midrule
GFHF      & ---      & ---    & --- & 0\%              & 0.9663$\pm$0.00482           & \textbf{0.95803$\pm$0.00749} & 0.94353$\pm$0.011            & 0.8691$\pm$0.03623           \\
GTAM      & ---      & 0.0101 & --- & 0\%              & 0.91417$\pm$0.01345          & 0.894$\pm$0.01539            & 0.89077$\pm$0.02639          & 0.81643$\pm$0.18266          \\
GTAM      & ---      & 99     & --- & 0\%              & 0.93$\pm$0.01825             & 0.85603$\pm$0.09293          & 0.77937$\pm$0.15401          & 0.7275$\pm$0.21115           \\
LGC       & 0.1      & ---    & --- & 0\%              & 0.95153$\pm$0.00814          & 0.931$\pm$0.00908            & 0.90177$\pm$0.01533          & 0.85363$\pm$0.04505          \\
LGC       & 0.9      & ---    & --- & 0\%              & 0.96397$\pm$0.00613          & 0.95267$\pm$0.00829          & 0.9332$\pm$0.01439           & 0.9028$\pm$0.03794           \\
LE      & ---      & ---    & 0.2 & 0\%              & \textbf{0.96847$\pm$0.00749} & 0.94923$\pm$0.01231          & \textbf{0.9446$\pm$0.01101}  & \textbf{0.9403$\pm$0.02737}  \\
\midrule
GFHF      & ---      & ---    & --- & 5\%              & \textbf{0.95873$\pm$0.00592} & \textbf{0.94947$\pm$0.0092}  & \textbf{0.932$\pm$0.01735}   & 0.8691$\pm$0.03623           \\
GTAM      & ---      & 0.0101 & --- & 5\%              & 0.91097$\pm$0.01546          & 0.88933$\pm$0.01185          & 0.85727$\pm$0.12625          & 0.81643$\pm$0.18266          \\
GTAM      & ---      & 99     & --- & 5\%              & 0.9207$\pm$0.0193            & 0.86437$\pm$0.04445          & 0.8209$\pm$0.11288           & 0.7275$\pm$0.21115           \\
LGC       & 0.1      & ---    & --- & 5\%              & 0.92447$\pm$0.00844          & 0.89643$\pm$0.01182          & 0.86153$\pm$0.0282           & 0.85363$\pm$0.04505          \\
LGC       & 0.9      & ---    & --- & 5\%              & 0.95573$\pm$0.00784          & 0.9402$\pm$0.01238           & 0.91723$\pm$0.01885          & 0.9028$\pm$0.03794           \\
LE      & ---      & ---    & 0.2 & 5\%              & 0.9569$\pm$0.0076            & 0.93263$\pm$0.01935          & 0.92637$\pm$0.01644          & \textbf{0.9403$\pm$0.02737}  \\ \midrule
GFHF      & ---      & ---    & --- & 10\%             & \textbf{0.94733$\pm$0.00925} & \textbf{0.938$\pm$0.01258}   & \textbf{0.92163$\pm$0.01562} & 0.7884$\pm$0.04509           \\
GTAM      & ---      & 0.0101 & --- & 10\%             & 0.9094$\pm$0.01513           & 0.8899$\pm$0.01756           & 0.88633$\pm$0.02636          & 0.79257$\pm$0.20503          \\
GTAM      & ---      & 99     & --- & 10\%             & 0.9076$\pm$0.0226            & 0.84683$\pm$0.05399          & 0.79057$\pm$0.14783          & 0.70057$\pm$0.21519          \\
LGC       & 0.1      & ---    & --- & 10\%             & 0.89287$\pm$0.01274          & 0.8549$\pm$0.02015           & 0.82667$\pm$0.03224          & 0.7574$\pm$0.04018           \\
LGC       & 0.9      & ---    & --- & 10\%             & 0.94417$\pm$0.01054          & 0.92523$\pm$0.01453          & 0.89633$\pm$0.02256          & 0.83467$\pm$0.04038          \\
LE      & ---      & ---    & 0.2 & 10\%             & 0.9328$\pm$0.01464           & 0.9056$\pm$0.03402           & 0.91053$\pm$0.02532          & \textbf{0.88707$\pm$0.05757} \\ \midrule
GFHF      & ---      & ---    & --- & 20\%             & \textbf{0.9167$\pm$0.0114}   & \textbf{0.90853$\pm$0.01657} & \textbf{0.86557$\pm$0.03994} & 0.74327$\pm$0.04684          \\
GTAM      & ---      & 0.0101 & --- & 20\%             & 0.9057$\pm$0.0167            & 0.8349$\pm$0.14307           & 0.84523$\pm$0.13535          & 0.74103$\pm$0.19152          \\
GTAM      & ---      & 99     & --- & 20\%             & 0.87423$\pm$0.07213          & 0.82197$\pm$0.07127          & 0.64137$\pm$0.21391          & 0.6952$\pm$0.1864            \\
LGC       & 0.1      & ---    & --- & 20\%             & 0.81407$\pm$0.01485          & 0.79053$\pm$0.02499          & 0.74847$\pm$0.03708          & 0.7162$\pm$0.05008           \\
LGC       & 0.9      & ---    & --- & 20\%             & 0.9156$\pm$0.0122            & 0.8854$\pm$0.01709           & 0.83363$\pm$0.04218          & 0.79637$\pm$0.05016          \\
LE      & ---      & ---    & 0.2 & 20\%             & 0.8742$\pm$0.01908           & 0.85657$\pm$0.03362          & 0.84883$\pm$0.05813          & \textbf{0.8678$\pm$0.07146}  \\ \midrule
GFHF      & ---      & ---    & --- & 35\%             & 0.7978$\pm$0.03554           & 0.80467$\pm$0.03405          & \textbf{0.72573$\pm$0.06141} & 0.6228$\pm$0.05523           \\
GTAM      & ---      & 0.0101 & --- & 35\%             & \textbf{0.87137$\pm$0.11513} & \textbf{0.88093$\pm$0.02184} & 0.65207$\pm$0.26244          & 0.6152$\pm$0.28324           \\
GTAM      & ---      & 99     & --- & 35\%             & 0.71907$\pm$0.15547          & 0.70397$\pm$0.1872           & 0.60963$\pm$0.21258          & 0.68193$\pm$0.18541          \\
LGC       & 0.1      & ---    & --- & 35\%             & 0.66907$\pm$0.02362          & 0.6662$\pm$0.02657           & 0.63407$\pm$0.04141          & 0.6185$\pm$0.06521           \\
LGC       & 0.9      & ---    & --- & 35\%             & 0.79673$\pm$0.03283          & 0.77063$\pm$0.02563          & 0.7005$\pm$0.05343           & 0.67513$\pm$0.07325          \\
LE      & ---      & ---    & 0.2 & 35\%             & 0.72327$\pm$0.03957          & 0.75027$\pm$0.04744          & 0.72067$\pm$0.08468          & \textbf{0.7747$\pm$0.11736}  \\ \bottomrule
\end{tabular}
\caption{Digit1 dataset: tolerance of classifiers to label noise given an affinity matrix based on a mutual KNN graph (k=15) and constant weights.}
\label{tab:digit1}
\end{table*}

\paragraph{COIL2}
The results for COIL2 are presented on Table \ref{tab:coil2} and the following are the main observation::
\begin{itemize}
    \item Configurations that emphasize label fitting (big $\alpha$, small $\mu$, also GFHF to some extent) were again susceptible to noise.
    \item As discussed before, there are some small clusters in the COIL2 dataset that may be left unlabeled by chance if we have few labels. Unsurprisingly, the performance of all algorithms is poor when only 1\% of the data is labeled.
    \item As opposed to the Digit1 dataset, the chosen configuration for the LE algorithm produced overall poor results for this dataset. 
\end{itemize}

\begin{table*}[!htb]
\scriptsize
\begin{tabular}{@{}lllllllll@{}}
\toprule
Algorithm & $\alpha$ & $\mu$  & p   & Noise & Acc. (10\% labeled)          & Acc. (5\% labeled)           & Acc. (2.5\% labeled)         & Acc. (1\% labeled)           \\ \midrule
GFHF      & ---      & ---    & --- & 0\%              & 0.87137$\pm$0.01113          & 0.84417$\pm$0.01818          & 0.7724$\pm$0.04148           & 0.64573$\pm$0.05772          \\
GTAM      & ---      & 0.0101 & --- & 0\%              & 0.6471$\pm$0.03938           & 0.64813$\pm$0.04815          & 0.62053$\pm$0.03334          & 0.57053$\pm$0.05646          \\
GTAM      & ---      & 99     & --- & 0\%              & 0.7399$\pm$0.04188           & 0.6727$\pm$0.05801           & 0.64077$\pm$0.06753          & 0.57143$\pm$0.06799          \\
LGC       & 0.1      & ---    & --- & 0\%              & \textbf{0.88227$\pm$0.0122}  & 0.8451$\pm$0.01494           & 0.77243$\pm$0.03421          & 0.65743$\pm$0.04943          \\
LGC       & 0.9      & ---    & --- & 0\%              & 0.87227$\pm$0.00997          & \textbf{0.85$\pm$0.01756}    & \textbf{0.78407$\pm$0.03603} & \textbf{0.66353$\pm$0.05605} \\
LGC       & 0.99     & ---    & --- & 0\%              & 0.8037$\pm$0.02014           & 0.79213$\pm$0.01875          & 0.76037$\pm$0.04172          & 0.65007$\pm$0.05907          \\
LE      & ---      & ---    & 0.2 & 0\%              & 0.77363$\pm$0.02066          & 0.7255$\pm$0.02641           & 0.67757$\pm$0.04761          & 0.53833$\pm$0.03236          \\ \midrule
GFHF      & ---      & ---    & --- & 5\%              & \textbf{0.86013$\pm$0.01477} & 0.82577$\pm$0.02436          & 0.7431$\pm$0.04876           & 0.64573$\pm$0.05772          \\
GTAM      & ---      & 0.0101 & --- & 5\%              & 0.64783$\pm$0.0386           & 0.64437$\pm$0.05689          & 0.61803$\pm$0.03749          & 0.57053$\pm$0.05646          \\
GTAM      & ---      & 99     & --- & 5\%              & 0.7287$\pm$0.04951           & 0.66817$\pm$0.06479          & 0.62053$\pm$0.06686          & 0.57143$\pm$0.06799          \\
LGC       & 0.1      & ---    & --- & 5\%              & 0.85753$\pm$0.01517          & 0.81307$\pm$0.02167          & 0.73607$\pm$0.04637          & 0.65743$\pm$0.04943          \\
LGC       & 0.9      & ---    & --- & 5\%              & 0.85857$\pm$0.01541          & \textbf{0.8288$\pm$0.02204}  & \textbf{0.7482$\pm$0.04705}  & \textbf{0.66353$\pm$0.05605} \\
LGC       & 0.99     & ---    & --- & 5\%              & 0.79603$\pm$0.01978          & 0.7861$\pm$0.02068           & 0.73557$\pm$0.04978          & 0.65007$\pm$0.05907          \\
LE      & ---      & ---    & 0.2 & 5\%              & 0.75607$\pm$0.02803          & 0.72067$\pm$0.03229          & 0.65623$\pm$0.06255          & 0.53833$\pm$0.03236          \\ \midrule
GFHF      & ---      & ---    & --- & 10\%             & 0.83827$\pm$0.02118          & 0.79743$\pm$0.03196          & 0.72053$\pm$0.05566          & 0.60763$\pm$0.05935          \\
GTAM      & ---      & 0.0101 & --- & 10\%             & 0.6394$\pm$0.0365            & 0.62237$\pm$0.06416          & 0.60077$\pm$0.04986          & 0.5455$\pm$0.06277           \\
GTAM      & ---      & 99     & --- & 10\%             & 0.70107$\pm$0.04853          & 0.65297$\pm$0.0662           & 0.6127$\pm$0.07432           & 0.54687$\pm$0.07252          \\
LGC       & 0.1      & ---    & --- & 10\%             & 0.82377$\pm$0.01817          & 0.7788$\pm$0.02341           & 0.70793$\pm$0.04371          & 0.61053$\pm$0.03887          \\
LGC       & 0.9      & ---    & --- & 10\%             & \textbf{0.83833$\pm$0.02073} & \textbf{0.80117$\pm$0.02539} & \textbf{0.72243$\pm$0.0493}  & \textbf{0.6215$\pm$0.05393}  \\
LGC       & 0.99     & ---    & --- & 10\%             & 0.78767$\pm$0.02221          & 0.76117$\pm$0.03398          & 0.71753$\pm$0.05787          & 0.6112$\pm$0.05916           \\
LE      & ---      & ---    & 0.2 & 10\%             & 0.742$\pm$0.03213            & 0.69963$\pm$0.03071          & 0.62883$\pm$0.05756          & 0.52583$\pm$0.03498          \\ \midrule
GFHF      & ---      & ---    & --- & 20\%             & 0.78103$\pm$0.03283          & 0.7382$\pm$0.04249           & 0.66477$\pm$0.06233          & 0.59247$\pm$0.03924          \\
GTAM      & ---      & 0.0101 & --- & 20\%             & 0.6218$\pm$0.04683           & 0.62197$\pm$0.05171          & 0.55543$\pm$0.06842          & 0.53377$\pm$0.06107          \\
GTAM      & ---      & 99     & --- & 20\%             & 0.6769$\pm$0.05176           & 0.6416$\pm$0.07097           & 0.57017$\pm$0.07439          & 0.52957$\pm$0.0731           \\
LGC       & 0.1      & ---    & --- & 20\%             & 0.7582$\pm$0.02008           & 0.71627$\pm$0.03943          & 0.65063$\pm$0.0433           & 0.59193$\pm$0.04117          \\
LGC       & 0.9      & ---    & --- & 20\%             & \textbf{0.78673$\pm$0.0321}  & \textbf{0.7411$\pm$0.03398}  & 0.66677$\pm$0.05501          & \textbf{0.60463$\pm$0.04721} \\
LGC       & 0.99     & ---    & --- & 20\%             & 0.75993$\pm$0.03674          & 0.71393$\pm$0.03938          & \textbf{0.67417$\pm$0.07127} & 0.60063$\pm$0.05188          \\
LE      & ---      & ---    & 0.2 & 20\%             & 0.7017$\pm$0.03505           & 0.6671$\pm$0.04531           & 0.59263$\pm$0.07276          & 0.52113$\pm$0.03125          \\ \midrule
GFHF      & ---      & ---    & --- & 35\%             & 0.669$\pm$0.03288            & \textbf{0.6287$\pm$0.05995}  & \textbf{0.604$\pm$0.07103}   & 0.52897$\pm$0.05549          \\
GTAM      & ---      & 0.0101 & --- & 35\%             & 0.58273$\pm$0.06572          & 0.56527$\pm$0.07798          & 0.5362$\pm$0.07146           & 0.51247$\pm$0.06028          \\
GTAM      & ---      & 99     & --- & 35\%             & 0.58063$\pm$0.05542          & 0.55917$\pm$0.08284          & 0.52503$\pm$0.08105          & 0.5257$\pm$0.07549           \\
LGC       & 0.1      & ---    & --- & 35\%             & 0.64343$\pm$0.01774          & 0.60513$\pm$0.04418          & 0.5782$\pm$0.05113           & 0.5358$\pm$0.04665           \\
LGC       & 0.9      & ---    & --- & 35\%             & \textbf{0.67803$\pm$0.02664} & 0.62597$\pm$0.05176          & 0.59143$\pm$0.06305          & \textbf{0.54127$\pm$0.05625} \\
LGC       & 0.99     & ---    & --- & 35\%             & 0.67267$\pm$0.05051          & 0.62133$\pm$0.0691           & 0.59773$\pm$0.07903          & 0.53153$\pm$0.05274          \\
LE      & ---      & ---    & 0.2 & 35\%             & 0.6257$\pm$0.03453           & 0.57627$\pm$0.05322          & 0.55373$\pm$0.07591          & 0.509$\pm$0.03506            \\ \bottomrule
\end{tabular}
\caption{COIL2 dataset: tolerance of classifiers to label noise given an affinity matrix based on a mutual KNN graph (k=15) and constant weights.}
\label{tab:coil2}
\end{table*}

\paragraph{g241c}
The results for g241c are presented on Table \ref{tab:g241c} and the following are the main observation:
\begin{itemize}
    \item LE, much like in Digit1, stayed extremely consistent when we reduced the amount of labeled instances. Moreover, it was also barely affected by label noise, up to 20\% of flipped labels.
    \item GTAM with $\mu = 99.0$ performed well when 10\% of instances were labeled.
\end{itemize}

\begin{table*}[!htb]
\scriptsize
\begin{tabular}{@{}lllllllll@{}}
\toprule
Algorithm & $\alpha$ & $\mu$  & p   & Noise & Acc. (10\% labeled)          & Acc. (5\% labeled)           & Acc. (2.5\% labeled)         & Acc. (1\% labeled)           \\ \midrule
GFHF      & ---      & ---    & --- & 0\%              & 0.70347$\pm$0.01052          & 0.6407$\pm$0.02062           & 0.5566$\pm$0.017             & 0.51153$\pm$0.00703          \\
GTAM      & ---      & 0.0101 & --- & 0\%              & 0.62193$\pm$0.02005          & 0.53773$\pm$0.02806          & 0.52107$\pm$0.01646          & 0.50827$\pm$0.00633          \\
GTAM      & ---      & 99     & --- & 0\%              & \textbf{0.73193$\pm$0.01415} & 0.68573$\pm$0.04224          & 0.64483$\pm$0.07759          & 0.59197$\pm$0.07209          \\
LGC       & 0.1      & ---    & --- & 0\%              & 0.67837$\pm$0.01376          & 0.648$\pm$0.01779            & 0.6197$\pm$0.02427           & 0.58137$\pm$0.03027          \\
LGC       & 0.9      & ---    & --- & 0\%              & 0.70367$\pm$0.00998          & 0.65757$\pm$0.0185           & 0.61397$\pm$0.02458          & 0.55093$\pm$0.0183           \\
LE      & ---      & ---    & 0.2 & 0\%              & 0.72677$\pm$0.01454          & \textbf{0.7103$\pm$0.02466}  & \textbf{0.71287$\pm$0.03458} & \textbf{0.71863$\pm$0.03359} \\ \midrule
GFHF      & ---      & ---    & --- & 5\%              & 0.6829$\pm$0.01458           & 0.62147$\pm$0.02377          & 0.5514$\pm$0.01516           & 0.51153$\pm$0.00703          \\
GTAM      & ---      & 0.0101 & --- & 5\%              & 0.61253$\pm$0.01803          & 0.5276$\pm$0.0268            & 0.52097$\pm$0.01423          & 0.50827$\pm$0.00633          \\
GTAM      & ---      & 99     & --- & 5\%              & \textbf{0.71687$\pm$0.0244}  & 0.6786$\pm$0.06005           & 0.628$\pm$0.07757            & 0.59197$\pm$0.07209          \\
LGC       & 0.1      & ---    & --- & 5\%              & 0.66003$\pm$0.01565          & 0.62997$\pm$0.01821          & 0.61053$\pm$0.01945          & 0.58137$\pm$0.03027          \\
LGC       & 0.9      & ---    & --- & 5\%              & 0.68393$\pm$0.0137           & 0.63987$\pm$0.01953          & 0.60533$\pm$0.02315          & 0.55093$\pm$0.0183           \\
LE      & ---      & ---    & 0.2 & 5\%              & 0.7102$\pm$0.0181            & \textbf{0.69447$\pm$0.03147} & \textbf{0.7078$\pm$0.03842}  & \textbf{0.71863$\pm$0.03359} \\ \midrule
GFHF      & ---      & ---    & --- & 10\%             & 0.6606$\pm$0.01638           & 0.6099$\pm$0.02241           & 0.54527$\pm$0.01623          & 0.5085$\pm$0.00537           \\
GTAM      & ---      & 0.0101 & --- & 10\%             & 0.60307$\pm$0.02075          & 0.5228$\pm$0.02499           & 0.52073$\pm$0.0141           & 0.5068$\pm$0.00535           \\
GTAM      & ---      & 99     & --- & 10\%             & \textbf{0.6945$\pm$0.03967}  & 0.6739$\pm$0.05051           & 0.62383$\pm$0.05826          & 0.55377$\pm$0.09596          \\
LGC       & 0.1      & ---    & --- & 10\%             & 0.64003$\pm$0.01571          & 0.61523$\pm$0.016            & 0.60057$\pm$0.02159          & 0.55857$\pm$0.02472          \\
LGC       & 0.9      & ---    & --- & 10\%             & 0.6607$\pm$0.0151            & 0.62523$\pm$0.01761          & 0.59517$\pm$0.02634          & 0.5338$\pm$0.0167            \\
LE      & ---      & ---    & 0.2 & 10\%             & 0.69297$\pm$0.02247          & \textbf{0.6839$\pm$0.03009}  & \textbf{0.69477$\pm$0.04599} & \textbf{0.68847$\pm$0.0692}  \\ \midrule
GFHF      & ---      & ---    & --- & 20\%             & 0.62837$\pm$0.01906          & 0.5866$\pm$0.02041           & 0.5283$\pm$0.01789           & 0.50633$\pm$0.00477          \\
GTAM      & ---      & 0.0101 & --- & 20\%             & 0.58323$\pm$0.02296          & 0.5188$\pm$0.01921           & 0.51103$\pm$0.01532          & 0.50337$\pm$0.00511          \\
GTAM      & ---      & 99     & --- & 20\%             & \textbf{0.66947$\pm$0.0404}  & 0.64673$\pm$0.06834          & 0.588$\pm$0.06831            & 0.54073$\pm$0.10164          \\
LGC       & 0.1      & ---    & --- & 20\%             & 0.60867$\pm$0.01468          & 0.58917$\pm$0.01729          & 0.56857$\pm$0.02967          & 0.55023$\pm$0.0246           \\
LGC       & 0.9      & ---    & --- & 20\%             & 0.6264$\pm$0.01554           & 0.5987$\pm$0.01866           & 0.56233$\pm$0.02683          & 0.52747$\pm$0.01601          \\
LE      & ---      & ---    & 0.2 & 20\%             & 0.66477$\pm$0.01658          & \textbf{0.6617$\pm$0.03463}  & \textbf{0.66187$\pm$0.05589} & \textbf{0.67577$\pm$0.07659} \\ \midrule
GFHF      & ---      & ---    & --- & 35\%             & 0.57197$\pm$0.02135          & 0.5474$\pm$0.02112           & 0.51377$\pm$0.01673          & 0.50313$\pm$0.00464          \\
GTAM      & ---      & 0.0101 & --- & 35\%             & 0.54833$\pm$0.0269           & 0.50613$\pm$0.01172          & 0.50357$\pm$0.00876          & 0.5021$\pm$0.00459           \\
GTAM      & ---      & 99     & --- & 35\%             & 0.59563$\pm$0.06997          & 0.57303$\pm$0.0898           & 0.5422$\pm$0.11096           & 0.4712$\pm$0.09492           \\
LGC       & 0.1      & ---    & --- & 35\%             & 0.5589$\pm$0.01421           & 0.5499$\pm$0.01873           & 0.53453$\pm$0.02396          & 0.5248$\pm$0.02435           \\
LGC       & 0.9      & ---    & --- & 35\%             & 0.5699$\pm$0.01632           & 0.55553$\pm$0.02016          & 0.52963$\pm$0.02339          & 0.51163$\pm$0.01296          \\
LE      & ---      & ---    & 0.2 & 35\%             & \textbf{0.59607$\pm$0.03144} & \textbf{0.6015$\pm$0.04439}  & \textbf{0.5955$\pm$0.06125}  & \textbf{0.55513$\pm$0.12361} \\ \bottomrule
\end{tabular}
\caption{g241c dataset: tolerance of classifiers to label noise given an affinity matrix based on a mutual KNN graph (k=15) and constant weights.}
\label{tab:g241c}
\end{table*}

\begin{table*}[!htb]
\scriptsize
\begin{tabular}{@{}lllllllll@{}}
\toprule
Algorithm & $\alpha$ & $\mu$  & p   & Noise & Acc. (10\% labeled)          & Acc. (5\% labeled)           & Acc. (2.5\% labeled)         & Acc. (1\% labeled)           \\ \midrule
GFHF      & ---      & ---    & --- & 0\%   & 0.72027$\pm$0.03377          & 0.63047$\pm$0.03544          & 0.5644$\pm$0.02081           & 0.51733$\pm$0.00625          \\
GTAM      & ---      & 0.0101 & --- & 0\%   & 0.63267$\pm$0.03763          & 0.55153$\pm$0.01977          & 0.521$\pm$0.01879            & 0.51093$\pm$0.00813          \\
GTAM      & ---      & 99     & --- & 0\%   & 0.60233$\pm$0.05529          & 0.56617$\pm$0.05592          & 0.5578$\pm$0.04968           & 0.5275$\pm$0.04079           \\
LGC       & 0.1      & ---    & --- & 0\%   & 0.7191$\pm$0.01704           & 0.6684$\pm$0.02325           & 0.6279$\pm$0.03135           & 0.5845$\pm$0.03127           \\
LGC       & 0.9      & ---    & --- & 0\%   & 0.72953$\pm$0.02708          & 0.64937$\pm$0.03831          & 0.5864$\pm$0.0398            & 0.54543$\pm$0.03008          \\
LGC       & 0.99     & ---    & --- & 0\%   & 0.71717$\pm$0.04016          & 0.59127$\pm$0.01496          & 0.54243$\pm$0.00807          & 0.5169$\pm$0.00608           \\
LE      & ---      & ---    & 0.2 & 0\%   & \textbf{0.80157$\pm$0.01736} & \textbf{0.7899$\pm$0.01841}  & \textbf{0.79437$\pm$0.02908} & \textbf{0.7965$\pm$0.04376}  \\ \midrule
GFHF      & ---      & ---    & --- & 5\%   & 0.69807$\pm$0.03048          & 0.615$\pm$0.02923            & 0.56313$\pm$0.01795          & 0.51733$\pm$0.00625          \\
GTAM      & ---      & 0.0101 & --- & 5\%   & 0.6152$\pm$0.03757           & 0.53933$\pm$0.02494          & 0.51997$\pm$0.01701          & 0.51093$\pm$0.00813          \\
GTAM      & ---      & 99     & --- & 5\%   & 0.59557$\pm$0.05161          & 0.56453$\pm$0.05274          & 0.54083$\pm$0.05129          & 0.5275$\pm$0.04079           \\
LGC       & 0.1      & ---    & --- & 5\%   & 0.69597$\pm$0.01452          & 0.65313$\pm$0.021            & 0.61977$\pm$0.02606          & 0.5845$\pm$0.03127           \\
LGC       & 0.9      & ---    & --- & 5\%   & 0.7086$\pm$0.02779           & 0.6326$\pm$0.03265           & 0.58023$\pm$0.03003          & 0.54543$\pm$0.03008          \\
LGC       & 0.99     & ---    & --- & 5\%   & 0.70033$\pm$0.03799          & 0.58373$\pm$0.01325          & 0.5372$\pm$0.00627           & 0.5169$\pm$0.00608           \\
LE      & ---      & ---    & 0.2 & 5\%   & \textbf{0.78323$\pm$0.01894} & \textbf{0.7666$\pm$0.02911}  & \textbf{0.77563$\pm$0.02961} & \textbf{0.7965$\pm$0.04376}  \\ \midrule
GFHF      & ---      & ---    & --- & 10\%  & 0.682$\pm$0.02978            & 0.59637$\pm$0.02682          & 0.54997$\pm$0.0194           & 0.50997$\pm$0.00862          \\
GTAM      & ---      & 0.0101 & --- & 10\%  & 0.6036$\pm$0.03897           & 0.53437$\pm$0.02274          & 0.51773$\pm$0.01113          & 0.5076$\pm$0.00729           \\
GTAM      & ---      & 99     & --- & 10\%  & 0.588$\pm$0.0499             & 0.56713$\pm$0.05119          & 0.54967$\pm$0.05348          & 0.52703$\pm$0.04768          \\
LGC       & 0.1      & ---    & --- & 10\%  & 0.67547$\pm$0.01306          & 0.63107$\pm$0.02331          & 0.6054$\pm$0.03025           & 0.56057$\pm$0.02719          \\
LGC       & 0.9      & ---    & --- & 10\%  & 0.6906$\pm$0.02395           & 0.6118$\pm$0.02976           & 0.57897$\pm$0.03524          & 0.53783$\pm$0.02938          \\
LGC       & 0.99     & ---    & --- & 10\%  & 0.68187$\pm$0.03049          & 0.57317$\pm$0.01329          & 0.5334$\pm$0.00795           & 0.5111$\pm$0.00538           \\
LE      & ---      & ---    & 0.2 & 10\%  & \textbf{0.7612$\pm$0.03012}  & \textbf{0.742$\pm$0.04075}   & \textbf{0.75657$\pm$0.04886} & \textbf{0.77553$\pm$0.06545} \\ \midrule
GFHF      & ---      & ---    & --- & 20\%  & 0.63$\pm$0.02352             & 0.5851$\pm$0.0277            & 0.5354$\pm$0.01305           & 0.51203$\pm$0.00767          \\
GTAM      & ---      & 0.0101 & --- & 20\%  & 0.57373$\pm$0.03371          & 0.54927$\pm$0.0324           & 0.51017$\pm$0.01323          & 0.50967$\pm$0.00727          \\
GTAM      & ---      & 99     & --- & 20\%  & 0.57437$\pm$0.04291          & 0.55553$\pm$0.0503           & 0.53927$\pm$0.03964          & 0.52863$\pm$0.04908          \\
LGC       & 0.1      & ---    & --- & 20\%  & 0.63313$\pm$0.01402          & 0.60503$\pm$0.02187          & 0.58253$\pm$0.02417          & 0.5528$\pm$0.02706           \\
LGC       & 0.9      & ---    & --- & 20\%  & 0.63747$\pm$0.02224          & 0.5958$\pm$0.02599           & 0.5646$\pm$0.02208           & 0.5442$\pm$0.02162           \\
LGC       & 0.99     & ---    & --- & 20\%  & 0.62777$\pm$0.02915          & 0.5666$\pm$0.0175            & 0.52377$\pm$0.00627          & 0.5106$\pm$0.00533           \\
LE      & ---      & ---    & 0.2 & 20\%  & \textbf{0.7171$\pm$0.03315}  & \textbf{0.70643$\pm$0.04704} & \textbf{0.70163$\pm$0.08498} & \textbf{0.75443$\pm$0.0623}  \\ \midrule
GFHF      & ---      & ---    & --- & 35\%  & 0.56807$\pm$0.02268          & 0.53693$\pm$0.01901          & 0.5325$\pm$0.02477           & 0.50623$\pm$0.00699          \\
GTAM      & ---      & 0.0101 & --- & 35\%  & 0.53893$\pm$0.02929          & 0.5$\pm$0.02279              & 0.513$\pm$0.01594            & 0.50437$\pm$0.00613          \\
GTAM      & ---      & 99     & --- & 35\%  & 0.53273$\pm$0.02541          & 0.52007$\pm$0.0418           & 0.53187$\pm$0.04307          & 0.51217$\pm$0.04208          \\
LGC       & 0.1      & ---    & --- & 35\%  & 0.56817$\pm$0.01583          & 0.55393$\pm$0.01701          & 0.5477$\pm$0.02461           & 0.5214$\pm$0.02002           \\
LGC       & 0.9      & ---    & --- & 35\%  & 0.56977$\pm$0.01692          & 0.545$\pm$0.01948            & 0.5466$\pm$0.02157           & 0.52337$\pm$0.02477          \\
LGC       & 0.99     & ---    & --- & 35\%  & 0.5641$\pm$0.01915           & 0.51977$\pm$0.01218          & 0.5159$\pm$0.00841           & 0.5055$\pm$0.00491           \\
LE      & ---      & ---    & 0.2 & 35\%  & \textbf{0.61623$\pm$0.03584} & \textbf{0.6192$\pm$0.05953}  & \textbf{0.62753$\pm$0.08866} & \textbf{0.6549$\pm$0.09849}  \\ \bottomrule
\end{tabular}
\caption{g241n dataset: tolerance of classifiers to label noise given an affinity matrix based on a mutual KNN graph (k=15) and constant weights.}
\label{tab:g241n}
\end{table*}

\paragraph{g241n}
The results for g241n are presented on Table \ref{tab:g241n} and the following are the main observation:
\begin{itemize}
    \item The Laplacian Eigenmaps (LE) algorithm achieved the overall best results for all configurations.
    \item When 10\% of instances were labeled, GTAM did not perform as well here as it did for g241c. 
    \item The setting of parameters $\alpha$ and $\mu$ did not seem to affect accuracy much. 
\end{itemize}

\subsection{Discussion}
\label{sec:disc}
The results of the experiments show us that the behavior of the selected GSSL classifiers depends heavily on the assumptions that are consistent with the dataset. For example, the Digit1 dataset is very consistent with the manifold assumption that the label propagation algorithms make use of. As expected, most algorithms perform well, even with few labels. Furthermore, lowering the importance of fitting the initial labels (as controlled by some hyperparameter) works well as a means to keep the accuracy from deteriorating. However, it must be said that, as one increases the label noise and decreases the number of available labels, eventually we should get to a point where it is unreasonable for the classifiers to be extremely accurate.

Experiments also yielded some surprising findings: first, the GTAM algorithm appeared to have high variance in some cases. This may possibly be attributed to the fact that it uses a greedy optimization procedure, although further investigation is required.  Furthermore, the LE algorithm was overall the best for g241c and g241n algorithms. We believe that this is due to the data being made up of a few high-dimensional clusters. When this happens, perhaps the best solution really is to identify the clusters, and associate it with the label that appears most frequently. The eigenfunction with smallest positive eigenvalue is commonly used for (unsupervised) graph cuts. Therefore, we believe that restricting ourselves to the smoothest eigenfunctions adheres to this cluster-then-label approach. For this reason, using the first eigenfunction to identify the two clusters in  g241c should be ideal. This also  explains how LE could possibly have better results with less labeled data, as we chose to determine $p$, the number of eigenfunctions, as a percentage of the labeled data. In spite of that, we remark that LE had poor performance when we tried to use weights based on a radial basis function. It could be necessary to look at the eigenfunctions for each matrix to understand this behavior.

\section{Conclusion and Future Work} \label{sec:conlusion-and-future-works}

In many situations, the data one has to work with is far from the ideal scenario. Obtaining the labels to train a supervised classifier is often too costly and time-consuming. As a result, one ends up with a lot of unlabeled data. Semi-supervised learning makes the most out of this situation by making use of the unlabeled data to enforce a specific prior belief. However, most of the standard SSL algorithms treat the initial labels as the absolute truth. In practical scenarios, labeling mistakes are common and must be addressed. 

In this work, we have shown that, if the dataset is consistent with the manifold assumption, classifiers based on label propagation will work well, especially when the hyperparameter determining the importance of fitting the initial labels is decreased. On the other hand, other approaches must be used if the data does not lie in a low-dimensional manifold. Our results have shown us that the Laplacian eigenmaps (LE) algorithm performed better than label propagation when the data came from high-dimensional clusters. Outright restricting eigenfunctions has the potential to be very effective at avoiding overfitting, but ultimately relies on a suitable criteria for the number of eigenfunctions. We have also seen that, when there are very few labels, it gets much harder to detect noisy instances properly. 
We have analyzed some classic graph-based SSL algorithms subject to label noise. We have also evaluated the Graph Transduction through Alternating Minimization (GTAM) algorithm, with its bivariate cost function.  Our initial results also motivate a further analysis of some algorithms, such as varying the parameters and investigating the eigenfunctions of the LE algorithm in order to obtain the most robust algorithm to label noise. Finally, we also aim to compare the previous classifiers with methods based on the $\ell_1$-norm. 

\section{Acknowledgments}
This study was financed in part by the Coordination of Superior Level Staff Improvement (CAPES) - Finance Code 001 and Sao Paulo Research Foundation (FAPESP) grants \#2018/15014-0 and \#2018/01722-3.


\bibliographystyle{ACM-Reference-Format}
\bibliography{sample-bibliography} 

\end{document}